\documentclass[nofootinbib]{revtex4}  

\usepackage{latexsym,amsmath,amssymb,amsbsy}
\usepackage{amsfonts}
\usepackage[utf8]{inputenc}
\usepackage{color}
\usepackage{subfigure}
\usepackage{graphicx}

\newcommand{\e}{\mathrm{e}}

\begin{document}  
  
\title{Projective simulation applied to the grid-world and the mountain-car problem}

\author{Alexey A. Melnikov}
    \affiliation{Institut f{\"u}r Theoretische Physik,
Universit{\"a}t Innsbruck, Technikerstra{\ss}e 25, A-6020 Innsbruck}
\affiliation{Institut f{\"u}r Quantenoptik und Quanteninformation der
\"Osterreichischen Akademie der Wissenschaften, Innsbruck, Austria}
    
  \author{Adi Makmal}%
    \affiliation{Institut f{\"u}r Theoretische Physik,
Universit{\"a}t Innsbruck, Technikerstra{\ss}e 25, A-6020 Innsbruck}
\affiliation{Institut f{\"u}r Quantenoptik und Quanteninformation der
\"Osterreichischen Akademie der Wissenschaften, Innsbruck, Austria}
    
 \author{Hans J. Briegel}%
    \affiliation{Institut f{\"u}r Theoretische Physik,
Universit{\"a}t Innsbruck, Technikerstra{\ss}e 25, A-6020 Innsbruck}
\affiliation{Institut f{\"u}r Quantenoptik und Quanteninformation der
\"Osterreichischen Akademie der Wissenschaften, Innsbruck, Austria}

\begin{abstract}
We study the model of projective simulation (PS) which is a novel approach to artificial intelligence (AI). Recently it was shown that the PS agent performs well in a number of simple task environments, also when compared to standard models of reinforcement learning (RL). In this paper we study the performance of the PS agent further in more complicated scenarios. To that end we chose two well-studied benchmarking problems, namely the ``grid-world'' and the ``mountain-car'' problem, which challenge the model with large and continuous input space. We compare the performance of the PS agent model with those of existing models and show that the PS agent exhibits competitive performance also in such scenarios.
\end{abstract}

 \maketitle

\section{Introduction}
Projective simulation (PS) provides a new approach for realizing an artificial intelligent (AI) agent, based on a stochastic processing of information~\cite{briegel2012projective}. The model, which can be naturally applied to reinforcement learning (RL) tasks \cite{russell1995artificial, barto1998reinforcement, RL_2012_book}, was tested in a recent paper~\cite{mautner2013projective} on a number of discrete toy-problems and was shown to perform well, also in comparison with the standard models of Q-learning \cite{russell1995artificial} and extended learning classifier systems \cite{Wilson95}. 

In this paper, we study the performance of the PS agent in more complicated scenarios. One particular type of real-world tasks is navigation, in which an agent has to find an optimal path to a target. Here we chose two canonical, well studied navigation tasks, namely the ``grid-world''~\cite{sutton1990integrated} and the ``mountain-car''~\cite{moore1990efficient} problem. The grid-world task is commonly used to examine the performance of AI approaches in handling large input space and delayed reward~\cite{sutton1990integrated, crook2003learning, wiering2012reinforcement}.  The mountain-car task presents an additional challenge, by imposing a continuous input space~\cite{singh1996reinforcement, sutton1996generalization, smart2000practical, rasmussen2003gaussian, jong2006kernel, whiteson2006evolutionary, heidrich2008variable, sutton2012dyna}.

The paper has the following structure: Section~\ref{sec:PSmodel} shortly describes the basic features of the PS model. Then, in Sections~\ref{sec:GridWorld} and~\ref{sec:MountainCar} we examine the performance of PS in the grid-world and mountain-car tasks, respectively. Last, Section~\ref{sec:Conclusion} summarizes the obtained results and concludes the paper.

\section{The PS model - Brief summary}
\label{sec:PSmodel}

\begin{figure}[h!]
   \begin{center}
   \begin{tabular}{c}
   \includegraphics[height=6cm]{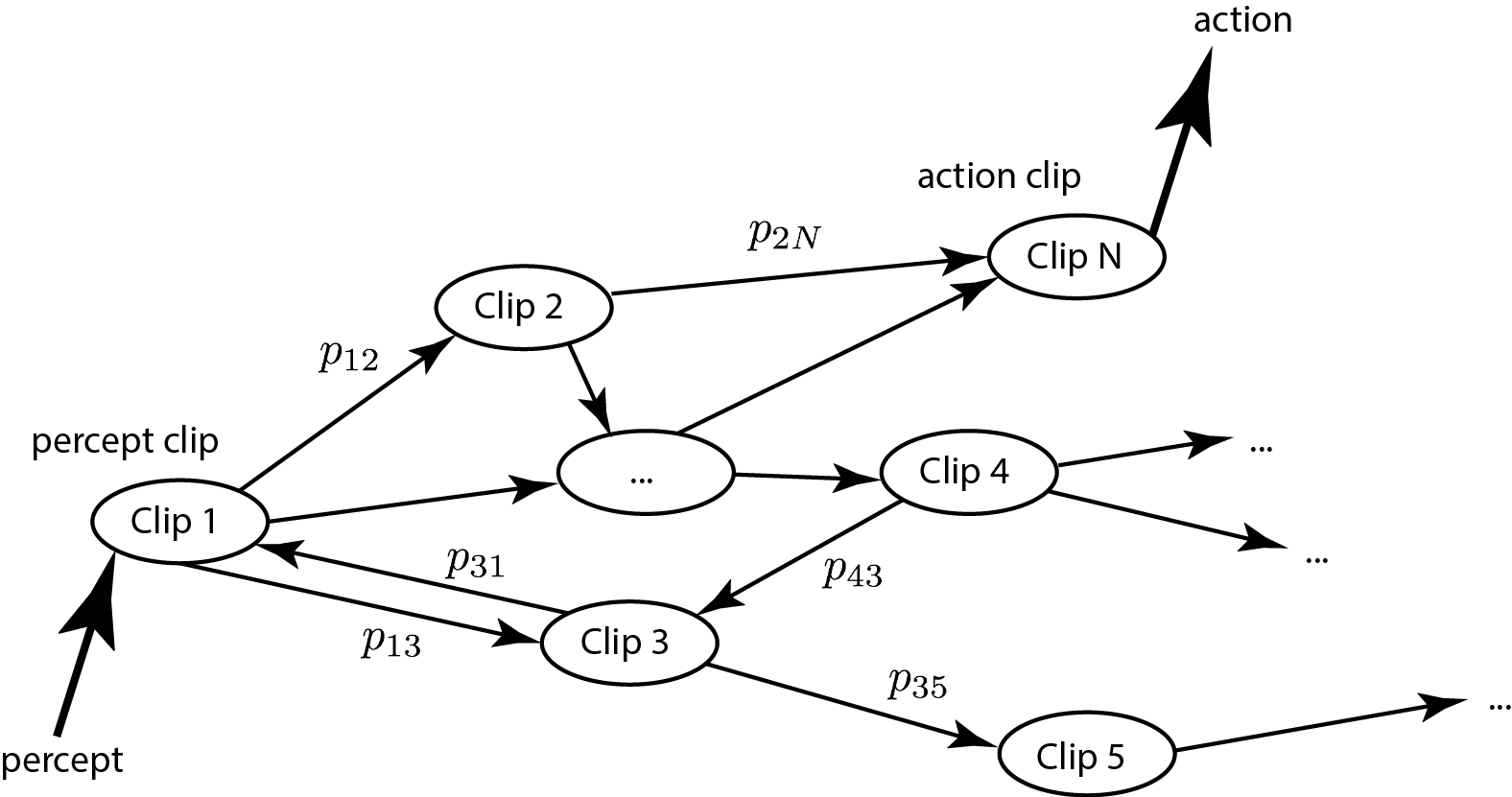}
   \end{tabular}
   \end{center}
\caption{The PS clip network. Arrows represent possible transitions between clips with conditional probabilities $p^{(t)}(c_j|c_i)$. The figure is adapted from~\cite{briegel2012projective}.}
\label{fig:clipNetwork}
   \end{figure}

For the benefit of the reader we first give a short description of the PS, for a more detailed description see~\cite{briegel2012projective, mautner2013projective}. The PS is an AI model in which the information received by the agent is processed in a so-called episodic $\&$ compositional memory (ECM). The ECM is described by a weighted network of ``clips" which are the units of the episodic memory: inputs lead to the excitation of corresponding percept clips, whereas the excitation of action clips triggers real action as output, as indicated in Fig.~\ref{fig:clipNetwork}. Once a percept-clip is excited, the excitation hops between clips probabilistically until it reaches an action-clip. In other words, in the PS agent, perceptual input and action output are connected through a random walk in the agent's memory. 

For illustration, let us consider a hypothetical PS network as shown in Fig.~\ref{fig:clipNetwork}. Each edge connects some clip $c_i$ with a clip $c_j$ and has a time-dependent weight $h^{(t)}(c_i,c_j)$ which we denote as $h$-value. The $h$-values represent the unnormalized strength of the edges, and determine the hopping probabilities from clip $c_i$ to clip $c_j$ according to
   \begin{equation}
  p^{(t)}(c_j|c_i) = \dfrac{h^{(t)}(c_i,c_j)}{\sum_{k} h^{(t)}(c_i,c_k)}.
\label{eq:probablititesBasic}
\end{equation}
In this paper, for the sake of comparison with existing models, we also consider an alternative expression for the hopping probability, known as the ``softmax" (or Boltzmann) distribution function
 \begin{equation}
  p^{(t)}(c_j|c_i) = \dfrac{\e^{\alpha' h^{(t)}(c_i,c_j)}}{\sum_{k} \e^{\alpha' h^{(t)}(c_i,c_k)}},
\label{eq:probablititesBoltzmann}
\end{equation}
where we always set $\alpha'=1$, unless stated otherwise. Note that using the  softmax expression merely rescales the hopping probabilities, such that a small difference in the $h$-values leads to a larger difference in the hopping probability. 

Initially, the $h$-values of all edges are set to $h^{(0)} = 1$, implying that no particular path in the clip network is preferred over any other, and hence that no action is more probable than the others. Then, as experience is built up, the clip-network is dynamically changed according to rewards perceived from the environment. Formally, at each time step, the $h$-values are updated as follows
\begin{equation}
  h^{(t+1)}(c_i,c_j) = h^{(t)}(c_i,c_j) - \gamma (h^{(t)}(c_i,c_j) - 1) + \lambda,
\label{eq:hupdate}
\end{equation}
where $0\leq\gamma\leq 1$ is a damping parameter and $\lambda$ is a non-negative reward given by the environment. Note that at each time step the weights of all edges are damped, but only the weights of those edges that were traversed in the very last random walk are increased by the $\lambda$ reward. This update rule allows the agent to learn through experience, in such a way that the probability to take rewarded actions is increased with time.

A useful generalization of the update rule of Eq.~(\ref{eq:hupdate}), denoted ``edge glow'', was added to the model in~\cite{mautner2013projective}. 
The ``edge glow" mechanism allows the agent to internally reward not only those edges (transitions) that were excited (taken) during the very last random walk, but also edges that were excited in previous time steps. This is realized by assigning to each edge of the PS network, apart from its weight, an additional time-dependent value $0\leq g \leq 1$, denoted as ``glow value'' and by using the modified update rule:  
\begin{equation}
  h^{(t+1)}(c_i,c_j) = h^{(t)}(c_i,c_j) - \gamma (h^{(t)}(c_i,c_j) - 1) + g^{(t)}(c_i,c_j)\lambda.
\label{eq:hupdate2}
\end{equation}
Each time an edge is visited, the corresponding $g$-value is set to $1$, following which it  
is decreased after each time step with a rate $\eta$:
\begin{equation}
  g^{(t+1)}(c_i,c_j) = g^{(t)}(c_i,c_j)(1 - \eta).
\label{eq:gupdate}
\end{equation}
The decay of the $g$ values ensures that the external reward has a different effect on edges that were excited at different time steps. In particular, edges that were excited in recent time steps are strengthen more than edges that were excited before, whose glow values $g^{(t)}$ have already decayed. Effectively, this means that percept-action pairs, that were experienced close to getting a reward, are more probable to reoccur than percept-action pairs that were encountered much before.   
The number of time steps that may pass between a random walk in the clip network and getting a reward such that the edges that were excited through that random walk are still strengthen by this reward, i.e.\ the number of time steps that are effectively ``remembered", is limited.
It is, however, controlled by the $\eta$ parameter through an inverse relation: the higher the value of $\eta$, the smaller the number of remembered steps. In particular, the agent remembers only the last few decisions (or, to be more precise, only the edges of last few random walks are strengthen) when $\eta \rightarrow 1$, and remembers almost every time step (that is, almost all previously excited edges are rewarded) when $\eta \rightarrow 0$. By setting $\eta=1$, only the last random walk path is rewarded and we revert back to the update rule of Eq.~(\ref{eq:gupdate}).

The generalized update rule of Eqs.~(\ref{eq:hupdate2})-(\ref{eq:gupdate}) enables the agent to correlate present rewards with previous actions and thereby to handle better with ``temporal correlation" scenarios, in which there exist a correlation between rewards and former actions. For example, this update rule was shown in~\cite{mautner2013projective} to be beneficial in scenarios in which the agent has to learn to refrain itself from actions that are instantly rewarded, at the benefit of obtaining a much larger reward for an action performed a few steps later, i.e.\ in scenarios where a greedy strategy is inferior. In this paper we further study the role of this update rule in scenarios in which a small reward is very much delayed.

\section{Grid World}
\label{sec:GridWorld}

\begin{figure}[h!]
   \begin{center}
   \begin{tabular}{c}
   \includegraphics[height=3.5cm]{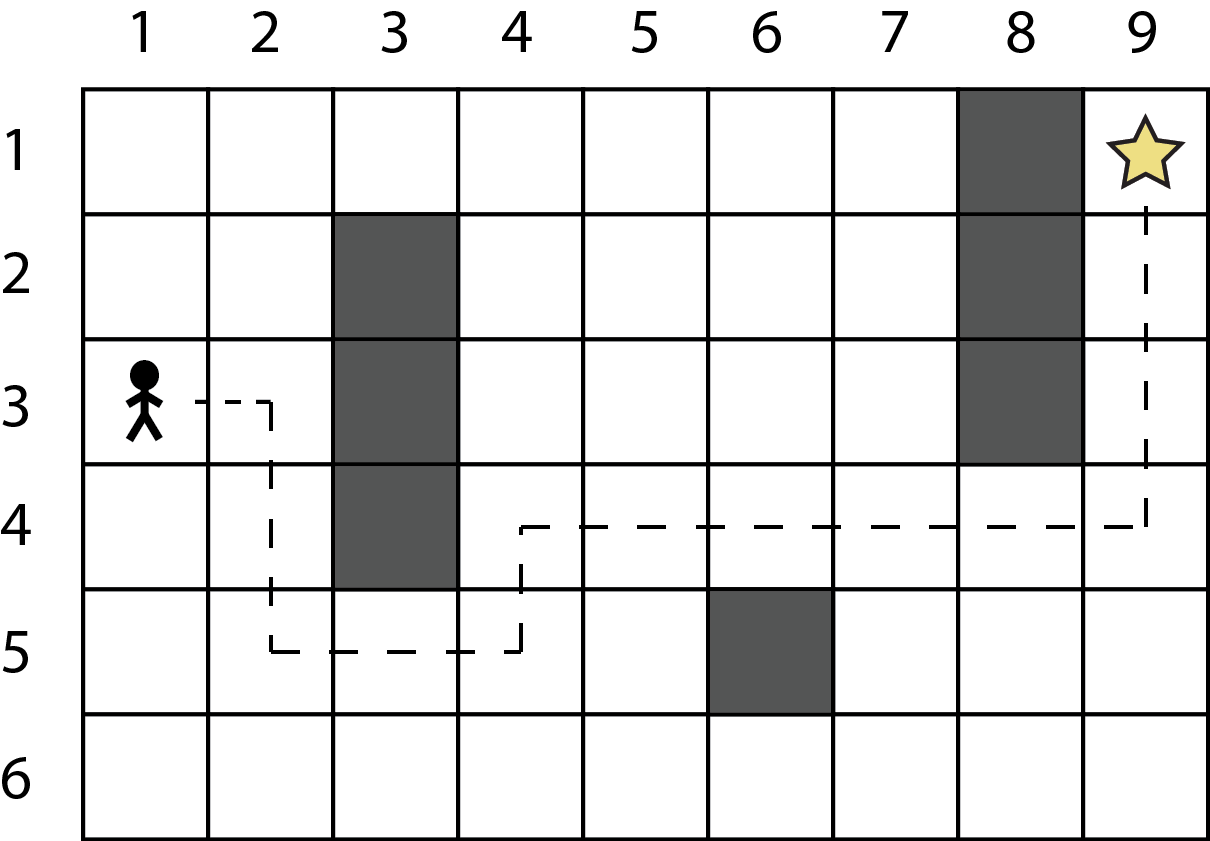}
   \end{tabular}
   \end{center}
\caption{The grid-world task: The goal of the game is to find the ``star''. At the beginning of each trial the agent is placed in the (1,3) cell, as shown. The shortest path to the goal is composed of 14 steps, one such optimal path is marked by a dashed line.}
\label{fig:GridWorldTask}
   \end{figure}

The grid-world environment~\cite{sutton1990integrated, crook2003learning} is a maze in which an agent should learn an optimal path to a fixed goal. The world is divided into discrete cells (or rooms) in which the agent can reside. At each time step the agent can move to one of the neighboring cells by choosing among four actions: left, right, up or down. Here we consider the maze from~\cite{sutton1990integrated} as shown in Fig.~\ref{fig:GridWorldTask}, which consists of 6 by 9 cells, some of which are walls (marked as black cells), and a goal (marked by a star) which is always located at the top right cell. At the beginning of each trial the agent is placed in the first cell of the third row from the top. If the agent decides to go to a square labeled as ``wall'' or to go beyond the grid, then no movement is performed but the time step is counted. The agent receives a reward of $\lambda = 1$ only after reaching the goal, which also marks the end of the trial. A performance of an agent in this task is evaluated by the number of steps it makes before reaching the goal at each trial. A learning agent will need less and less steps as it goes through more and more trials. The learning time can be defined by the number of trials required to get a certain level of performance. A more efficient agent will require a smaller number of trials to attain a substantial improvement of its performance. 

The main challenges posed by the grid-world task are its relative large input space (46 possible positions in our case) and the fact that the reward is much delayed. 
In fact, at the first trial, and for many time steps, the agent has no preference toward any direction until the goal is found by sheer coincidence. Only after the agent is rewarded for the first time, it can start developing a preference toward reaching the goal.

\begin{figure}[h!]
   \begin{center}
   \begin{tabular}{c}
   \includegraphics[height=4.5cm]{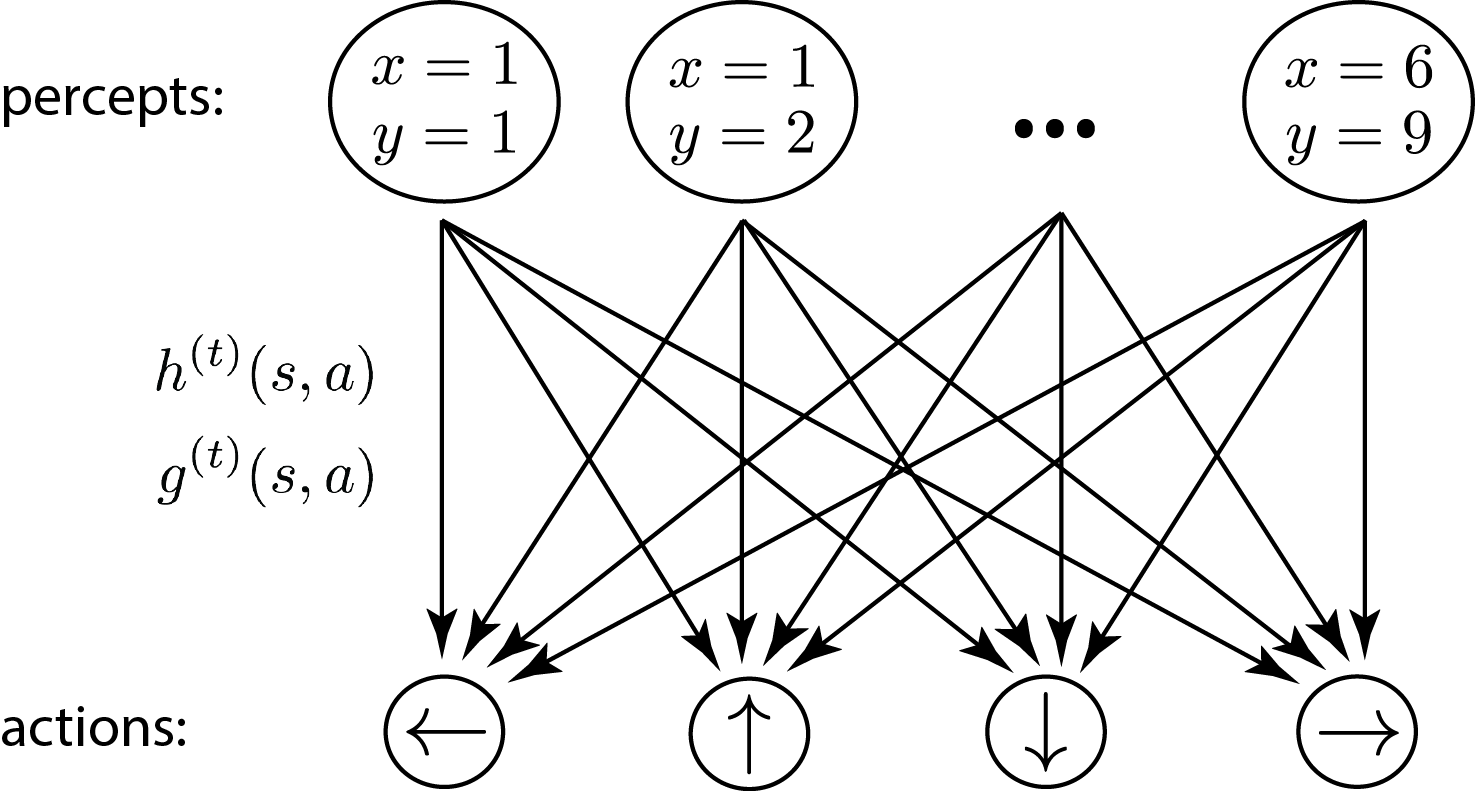}
   \end{tabular}
   \end{center}
\caption{The PS clip network in the grid-world task. First and second rows depict percept and action clips, respectively, and a directed edge leads from every percept to every action clip. Input is perceived as coordinates on the maze: a row number $x$ and a column number $y$. The network has 46 percept clips and 4 action clips. Each edge is associated with an $h$-value $h^{(t)}(s,a)$ and a glow value $g^{(t)}(s,a)$.}
\label{fig:memoryGrid}
   \end{figure}

In the following we examine the performance of the PS agent in the grid-world task.
To that end we use a two-layered clip network structure, as shown in Fig.~\ref{fig:memoryGrid}, composed of 46 percept-clips (first row in Fig.~\ref{fig:memoryGrid}) representing potential positions on the maze, 4 action-clips (second row in Fig.~\ref{fig:memoryGrid}) and directed edges connecting percepts (s) to actions (a). Each edge (s,a) between percept and action is assigned a time dependent $h$-value $h^{(t)}(s,a)$ and a glowing value $g^{(t)}(s,a)$, as explained in Sec.~\ref{sec:PSmodel}. Those values are then updated through experience, according to generalized update rules of Eqs.~(\ref{eq:hupdate2})-(\ref{eq:gupdate}). 
To obtain statistically meaningful results we average the PS performance over $10^4$ agents. (see \cite{mautner2013projective} for an error bars analysis of the PS agent's learning curves, albeit in a different scenario). 
 
\begin{figure}[h!]
    \centering
    \subfigure[]
    {
        \includegraphics[width=7cm]{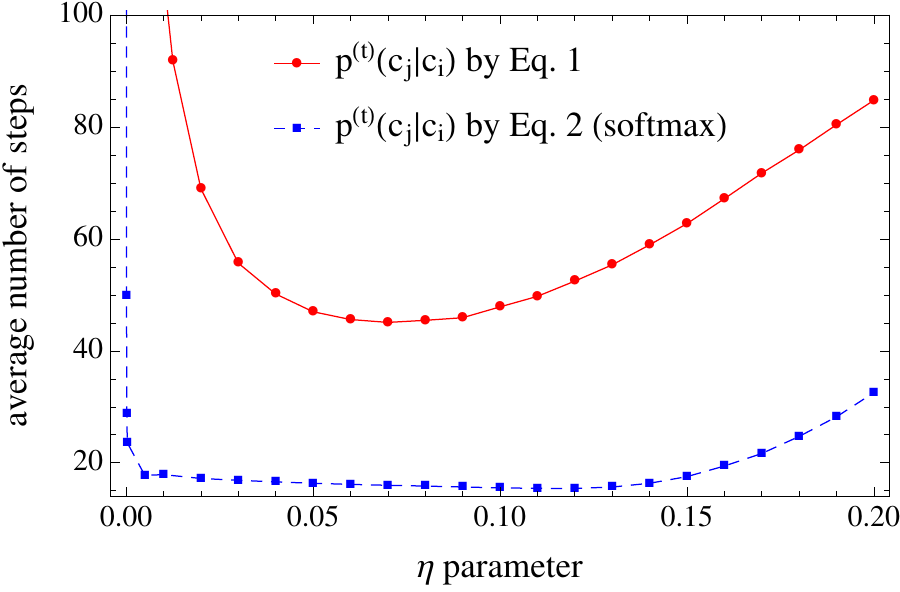}
        \label{fig:GridEta}
    }
    \subfigure[]
    {
        \includegraphics[width=7cm]{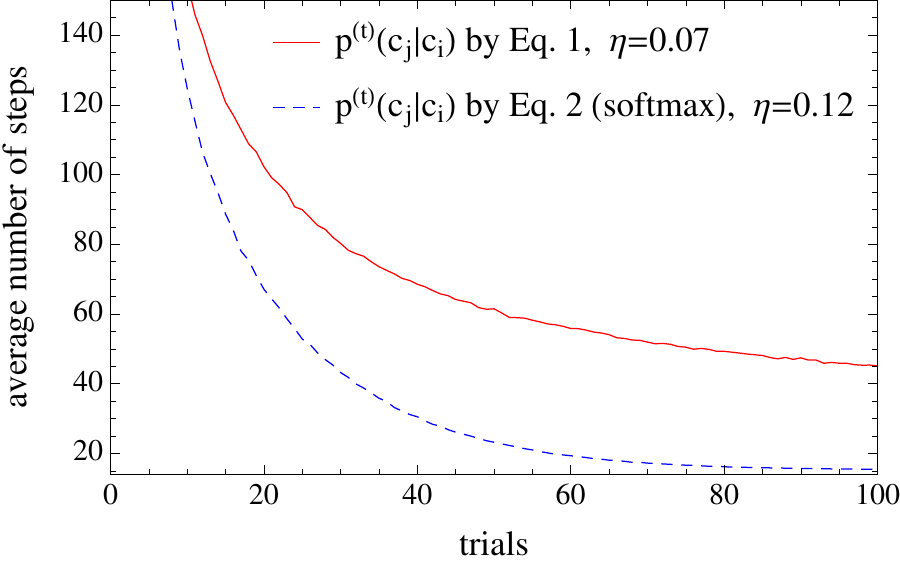}
        \label{fig:Gridlearning}
    }
    \caption{
    The performance of the PS agents in the grid-world task after 100 trials. Solid red curves depict the PS performance using the basic transition probability function (Eq.~(\ref{eq:probablititesBasic})). Dashed blue curves depict the PS performance using the softmax transition probability function (Eq.~(\ref{eq:probablititesBoltzmann})). A damping value of $\gamma=0$ is used throughout. All curves are averaged over $10^4$ agents. 
(a) The dependence of the PS performance on the $\eta$ parameter is shown after 100 trials. (b) PS learning curves are shown for optimal values of $\eta=0.07$ and $\eta=0.12$ (for 100 trials). 
The performance improves with the number of trials: from about 870 steps at the first trial to 45 (solid red) and 15.4 (dashed blue) steps, after 100 trials.}
    \label{fig:GridPerformance}
\end{figure}

As shown in previous works~\cite{briegel2012projective, mautner2013projective} 
the PS performance depends on the value of its internal $\gamma$ and $\eta$ damping parameters.
In particular, it was shown that a nonzero damping parameter $\gamma$, i.e.\ an ongoing process of forgetting, is beneficial when the environment changes, whereas for constant environments it merely limits the maximum achievable success probability of the agent. Since in the grid-world task the environment is constant we set $\gamma=0$ to avoid forgetting and to observe the model's best performance. The dependence of the PS performance on the value of the glow-damping parameter $\eta$ is, however, more involved. Fig.~\ref{fig:GridEta} shows the PS performance, characterized by the number of steps required to find the goal after 100 trials, as a function of the $\eta$ parameter. We consider both the basic and the softmax probability functions $p^{(t)}(c_j|c_i)$ as given in Eqs.~(\ref{eq:probablititesBasic}) and (\ref{eq:probablititesBoltzmann}), shown in solid red and dashed blue lines, respectively. 
One can see that in both cases the PS agent performs quite badly with $\eta\rightarrow 0$: even after 100 trials it requires more than 100 steps to reach the goal (in fact for $\eta=0$ the agents require 842 and 570 steps, after 100 trials, using the basic and the softmax functions, respectively, not shown). This is because a small $\eta$ parameter inhibits the decay of the edge glow, so that all previous actions are always rewarded with the same value $\lambda=1$. Since the PS agent starts the first trial with no preferred actions, the first path to the goal consists of completely random moves and is thus very long on average.  The probability of taking again the same long path of random moves increases and makes it almost impossible to learn something. Setting $\eta=1$ may even be worse. The reason is that with  $\eta=1$ all $g$ values are damped to $0$ and the ``edge glow'' mechanism is effectively turned off, such that only the last action in the trial can be learned. Setting an intermediate value can help as it allows to reward moves which are near the goal higher than those which are far from it. 
In other words, the last few actions before reaching the goal are  highly rewarded, whereas unwanted random moves at the beginning of the trial are less rewarded, if at all. From Fig.~\ref{fig:GridEta} one can see that there exists an optimal $\eta \approx 0.07$ for the PS agent with the basic transition probabilities of Eq.~(\ref{eq:probablititesBasic}) (solid red curve). This is in agreement with the findings of~\cite{mautner2013projective} in which an optimum $\eta$ value was also shown to exist for a different ``temporal correlation" scenario. Using the softmax function to define the transition probabilities (dashed blue curve in Fig.~\ref{fig:GridEta}) leads to an improvement in the agents performance. This improvement makes sense because with the softmax function even a small reward is enough to establish a high probability to repeat the same action. Moreover, it is seen that when the softmax probability function is used, the resulting performance (i.e.\ after 100 trials) is more robust against changes of $\eta$. In particular, there is a larger region of $\eta$ values for which the performance is nearly the best.

Fig.~\ref{fig:Gridlearning} shows the average performance of the PS agent as a function of the number of trials, using the optimal values of $\gamma=0$, $\eta=0.07$ (for the basic transition function, shown in solid red curve) and $\eta=0.12$ (for the softmax function, shown in dashed blue curve) as explained above. It is seen that as the number of trials increases, the PS agents find the goal in fewer and fewer steps on average, implying that the PS model is capable of learning in the grid-world environment.
In particular, after 100 trials it is seen that on average the PS agents reach the goal in about 45 steps using the basic transition function and in about 15.4 steps when using the softmax probability function. It is also seen that the initial learning rate, as captured by the initial slope of the learning curves is greater when using the softmax function. 

\begin{figure}[h!]
    \centering
    \subfigure[]
    {
        \includegraphics[width=7cm]{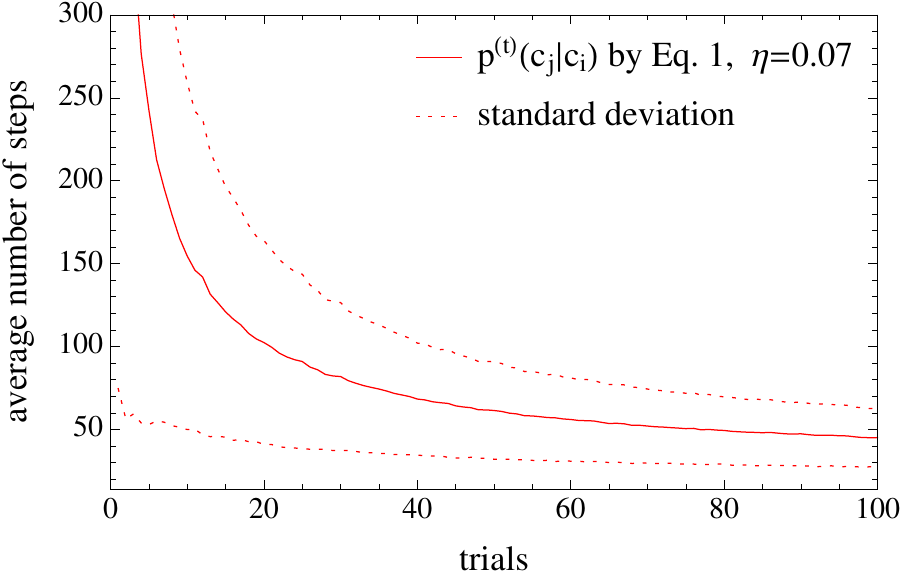}
        \label{fig:GridlearningSigma1}
    }
    \subfigure[]
    {
        \includegraphics[width=7cm]{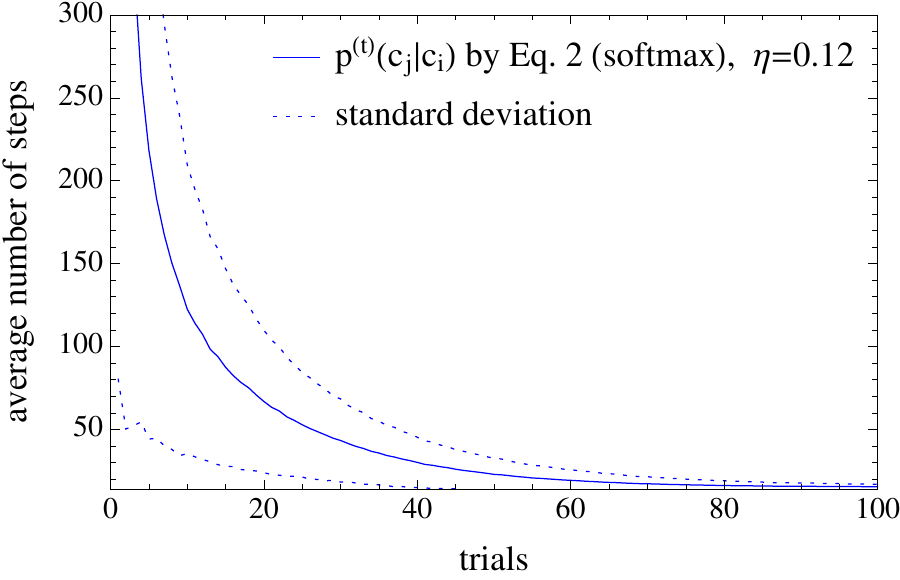}
        \label{fig:GridlearningSigma2}
    }
    \caption{The performance of the PS agents in the grid-world task during 100 trials, including standard deviations. The solid curves mark the averaged performance over $10^4$ agents. Dotted curves correspond to adding and subtracting one standard deviation $\sigma$ from the average value. 
(a) Using the basic transition probability of Eq.~(\ref{eq:probablititesBasic}), in red. (b) Using the softmax transition probability of Eq.~(\ref{eq:probablititesBoltzmann}), in blue.}
    \label{fig:GridPerformanceSigma}
\end{figure}

The performance of each individual agent may differ from the average performance. In order to check how much a single agent's behavior may deviate from the one of its fellow agents, i.e.\ differ from the average, we further computed the standard deviation ($\sigma$) of the performance for each trial. 
Fig.~\ref{fig:GridPerformanceSigma} shows the averaged performance of the PS agents in the grid-world (solid curves), plotted in the center of a $2\sigma$ envelope (dotted curves). It is seen that at the beginning the agent's performance varies to a large extent, and that as the number of trials increases, the agent's performance converges into a small region. In other words, the standard deviation decreases with experience and the agents are expected to perform more and more alike and as suggested by the average of their performance.
 
\begin{figure}[h!]
    \centering
    \subfigure[]
    {
        \includegraphics[width=7cm]{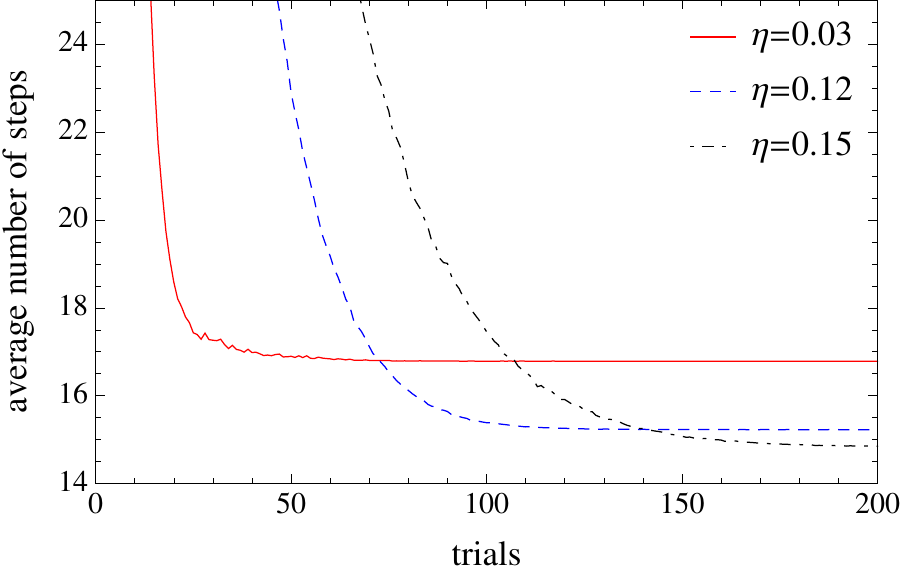}
        \label{fig:BestPerformance1}
    }
    \subfigure[]
    {
        \includegraphics[width=7cm]{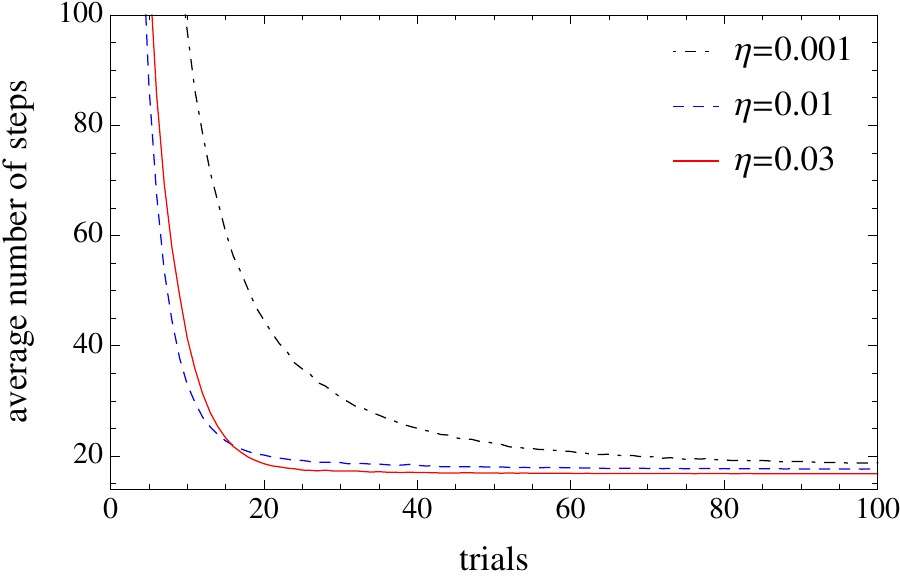}
        \label{fig:BestPerformance2}
    }
    \caption{The learning curves of the PS agent in the grid-world task, with different $\eta$ values. All curves are calculated using the softmax function with $\alpha'=1$. A damping value of $\gamma=0$ is used throughout. All curves are averaged over $10^4$ agents.
(a) A trade-off is observed between the best performance and the number of trials required to reach it. In particular, as $\eta$ increases the initial slope decreases (and more trials are needed to reach the best performance), yet a better performance is reached after the 200$^{\text{th}}$ trial. 
(b) Among the three $\eta$ values, the intermediate value of $\eta=0.01$ exhibits the largest initial slope.}
    \label{fig:GridBestPerformance1}
\end{figure}

So far, we have mostly concentrated on the performance of the agent in terms of the length of the chosen path. We saw that after 100 trials, different $\eta$ values result with different performances, and that there exist an optimal $\eta$ value for which this performance is the best. Next we also consider the effect of changing the value of the  $\eta$ parameter on the initial speed of the learning. 
Fig.~\ref{fig:BestPerformance1} shows three learning curves corresponding to three different values of $\eta=0.03, 0.12$ and $0.15$, in solid red, dashed blue and dash-dotted black, respectively. 
It is seen that as the value of $\eta$ increases, the initial slope of the learning curve decreases and a better performance is reached. This illustrates that the choice of an optimal $\eta$ depends on the number of trials the agent is given. In particular, it is seen that after 100 trials the dashed blue curve of $\eta=0.12$ exhibits the best performance, but that with $\eta=0.15$ (dash-dotted black curve) one can reach better performance after 150 trials. Moreover Fig.~\ref{fig:BestPerformance1} illustrates that for a finite number of trials there is a trade-off between learning speed and performance. In particular the solid red curve of $\eta=0.03$ shows that by slightly compromising on performance (with a path of less than 17 steps) a much faster learning is obtained. In general, by increasing the $\eta$ value further one can achieve better performance, but more learning trials are needed. The choice of $\eta$ can then be made according to the required property: performance versus speed. For completeness, we remark that by increasing the value of $\alpha'$, i.e.\ the exponent power of the softmax function of Eq.~(\ref{eq:probablititesBoltzmann}) one can improve on both the initial learning slope and the performance (not shown).

Next we ask, is there a limit for the initial learning speed or is there an optimum $\eta$ value for which the initial slope is the largest? Fig.~\ref{fig:BestPerformance2} shows that among $\eta=0.001, 0.01$ and $0.03$ in dash-dotted black, dashed blue and solid red curves, respectively, the largest initial slope is obtained by the intermediate value of $\eta=0.01$. This shows that an optimal $\eta$ value does exist, also with respect to the achievable initial slope. 

We now turn to evaluate the quality of the PS performance. To that end we compare the PS performance to the performance of the policy iteration (PI) model as reported in~\cite{sutton1990integrated} where the exact same rules were employed.
In particular, we compare the PS learning curve depicted in Fig.~\ref{fig:Gridlearning} to the learning curve of the PI agent as shown in Fig.\ 3 of Ref.~\cite{sutton1990integrated}.
The PI is a trial-and-error learning system. The policy, which is adjusted by a temporal-difference learning process, dictates which action is performed at any input state. It is realized as a table of values for each pair of input and action, and the PI agent takes an action stochastically using values from this table according to the softmax function. To allow a meaningful comparison we compare the performance of the basic PS model with those of the basic PI model, i.e.\ without any Dyna-style planning steps (or, equivalently, with no hypothetical experience, denoted as $k=0$ in Ref.~\cite{sutton1990integrated}). Moreover, since the PI employs the softmax policy, we compare it with the softmax curve of the PS shown in dashed blue in Fig.~\ref{fig:Gridlearning}. It is seen in Fig. 3 of Ref.~\cite{sutton1990integrated} that after about 80 trials the PI method (with no Dyna-style planning) reaches the goal in about 14 steps, i.e.\ roughly within the optimal number of steps. In comparison, the PS agents with the softmax function require on average 15.4 steps. These results are also summarized in Table \ref{table:grid_world_comparison}.

\begin{table}[h!]
\begin{tabular}{c|c|c}
  Model &$\#$Steps to goal after 100 trials & Parameters  \\[1mm] 
 \hline 
	PS & 45 & $\lambda = 1 $, $ \eta = 0.07 $, $ \gamma = 0 $ \\
    PS softmax & 15.4 & $\lambda = 1 $, $ \eta = 0.12 $, $ \gamma = 0 $ \\
\hline
	PI~\cite{sutton1990integrated}& 14 & $\beta=0.1, \gamma=0.9, \alpha=1000$\\
\end{tabular}
\caption{Grid world: performances of the PS model in comparison with the PI model, as reported in~\cite{sutton1990integrated}.}
\label{table:grid_world_comparison}
\end{table}

\section{Mountain car}
\label{sec:MountainCar}
\begin{figure}[h!]
   \begin{center}
   \begin{tabular}{c}
   \includegraphics[height=4.5cm]{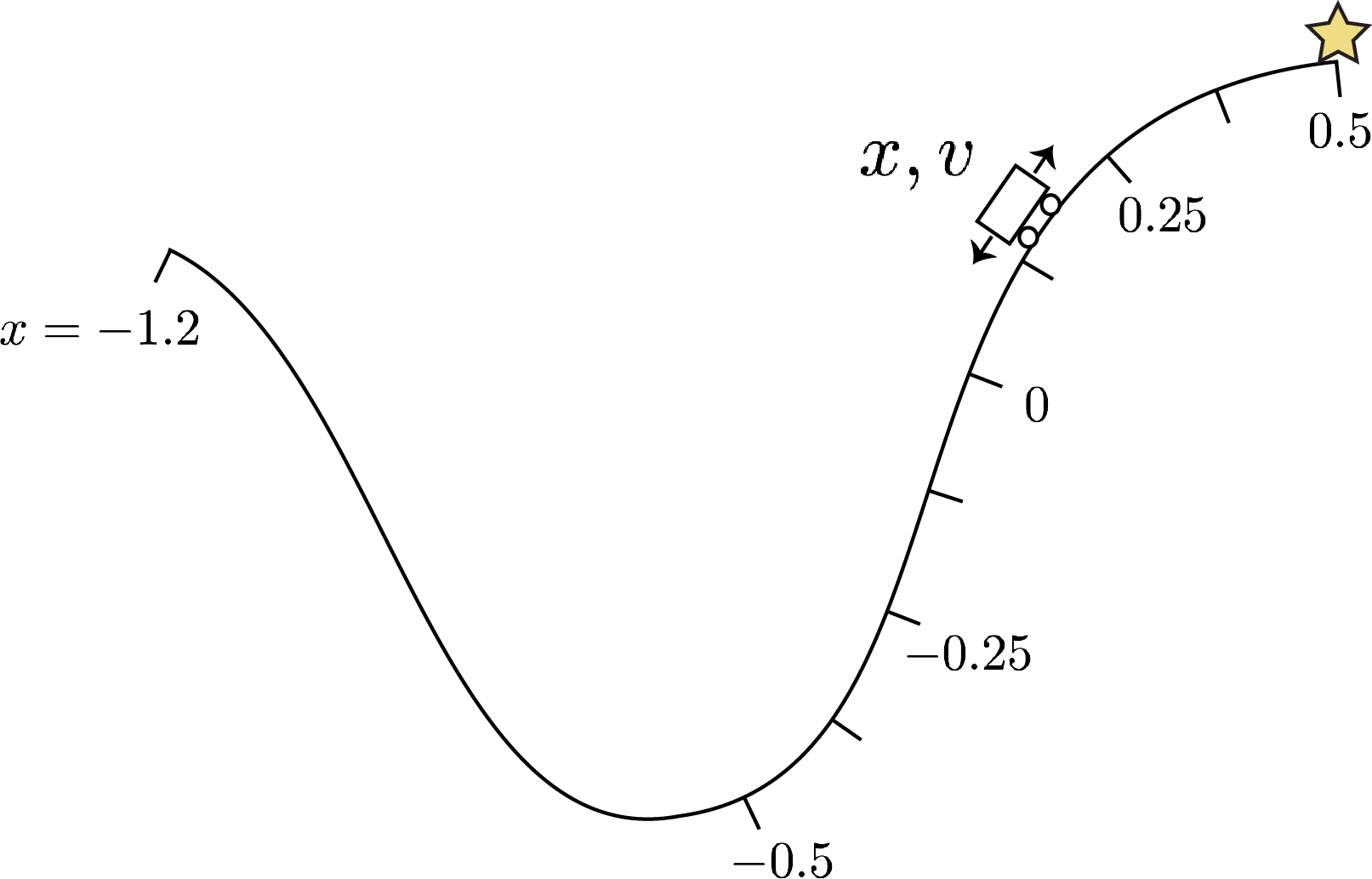}
   \end{tabular}
   \end{center}
\caption{The mountain-car task, a schematic drawing: the goal is to find the ``star'' at $x=0.5$. Marks on the road represent the $x$ coordinate.}
\label{fig:MountainTask}
   \end{figure}

In the mountain-car task, defined in~\cite{singh1996reinforcement, sutton1996generalization}, an agent drives a car on a surface between two hills, where a goal awaits at the top of the right hill, as shown in Fig.~\ref{fig:MountainTask}. At the beginning of each trial the agent has a random position $x_0$ and a random velocity $v_0$. Then, at each of the following time steps the agent receives its new position $x_\mathrm{new}$ and new velocity $v_\mathrm{new}$ as input and has to choose between three possible actions: forward thrust (to the right), no thrust, and reverse thrust (to the left). 
Once the agent finds the goal it is rewarded with $\lambda=1$ and the trial ends. Until then, like in the grid-world, the agent receives no rewards.  
To measure the performance of the agent we count the number of steps it requires to find the goal at each trial. Clearly, a well performed agent would require less steps as the number of trials increases. 

Here we follow the mountain-car rules as specified in~\cite{singh1996reinforcement}. In particular, the next $(x_\mathrm{new},v_\mathrm{new})$ coordinates are determined by the agent's own action and by the effect of gravity, according to 
\begin{equation}
  \begin{array}{ccl}
    v_\mathrm{new} & = & v_\mathrm{old} + 0.001*\mathrm{Action} - 0.0025\cos (3x_\mathrm{old}), \quad\mathrm{Action}\in\{-1,0,1\}\\
    x_\mathrm{new} & = & x_\mathrm{old} + v_\mathrm{old}.
  \end{array}
\label{eq:stateUpdate}
\end{equation}
The position of the car is bounded inside $-1.2 \leq x \leq 0.5$ and the goal is always placed at $x=0.5$. Trying to go beyond these bounds leaves the car on the boundary with zero velocity as if the car hit a wall (for the positive boundary this would simply result with reaching the goal). The velocity is similarly bounded inside  $-0.07 \leq v \leq 0.07$, i.e.\ its absolute value is reset to 0.07 if this value is exceeded \cite{singh1996reinforcement}. 

The mountain-car task is quite challenging: first, like in the grid-world, the reward is delayed and the agent has initially no information about the goal or its mere existence. It therefore has to move around randomly until it accidentally hits the goal and is finally rewarded; Second, unlike the grid-world scenario, an optimal path is not apparent. This is because in general it will not by sufficient to push a car directly to the goal, since its engine power is not strong enough and it will eventually roll back down. The agent would therefore need to drive the car back and forth to obtain sufficient potential energy. Appendix~\ref{sec:Appendix} provides a detailed analytical treatment of the physics that lays behind the game, and calculates an upper bound (though not necessarily tight) for the minimum number of required steps; Last but not least, the mountain-car task introduces a two-dimensional continuous input space, thereby confronting the agent with infinite number of possible input states.

   \begin{figure}[h!]
   \begin{center}
   \begin{tabular}{c}
   \includegraphics[height=4.5cm]{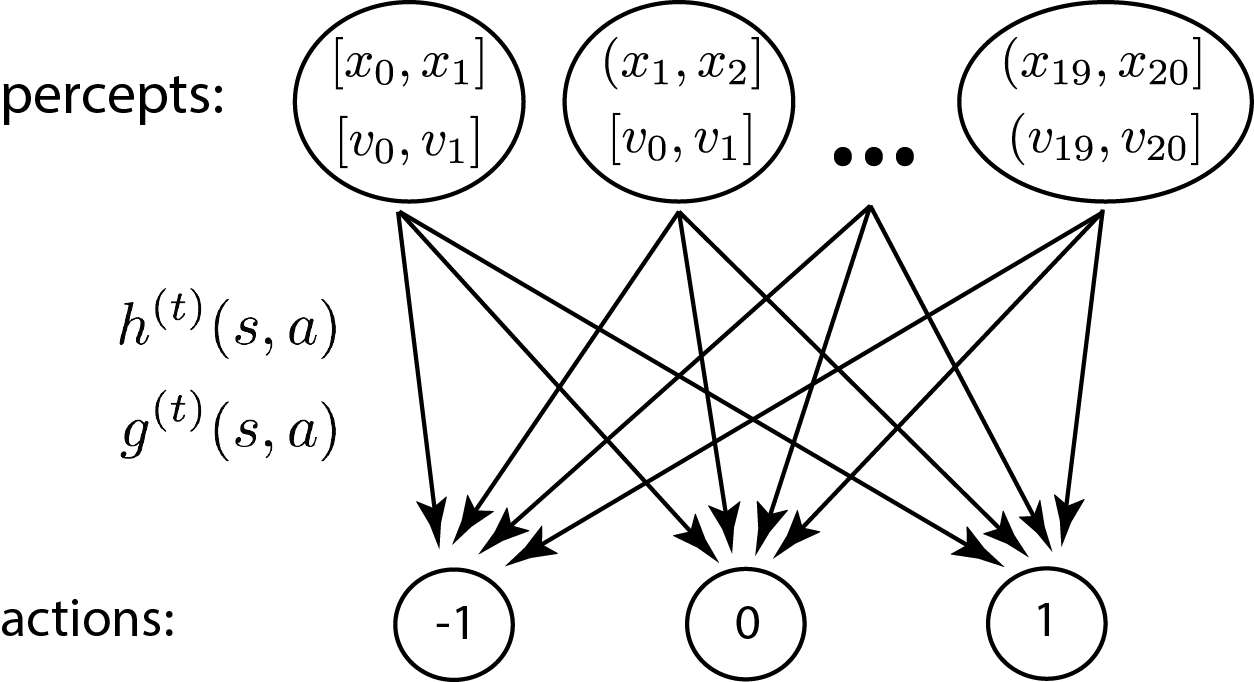}
   \end{tabular}
   \end{center}
\caption{The PS clip network in the mountain-car problem. 
First and second rows depict percept and action clips, respectively, and a directed edge leads from every percept to every action clip.
The continuous input space is discretized to 
a grid of uniform $(x,v)$ regions, each corresponds to a different possible input state. Here we used a PS network composed of 20 by 20 percept clips and 3 action clips. 
Each edge is associated with an $h$-value $h^{(t)}(s,a)$ and a glow value $g^{(t)}(s,a)$.}
\label{fig:memoryMC}
   \end{figure} 

\begin{figure}[h!]
    \centering
    \subfigure[]
    {
        \includegraphics[width=7cm]{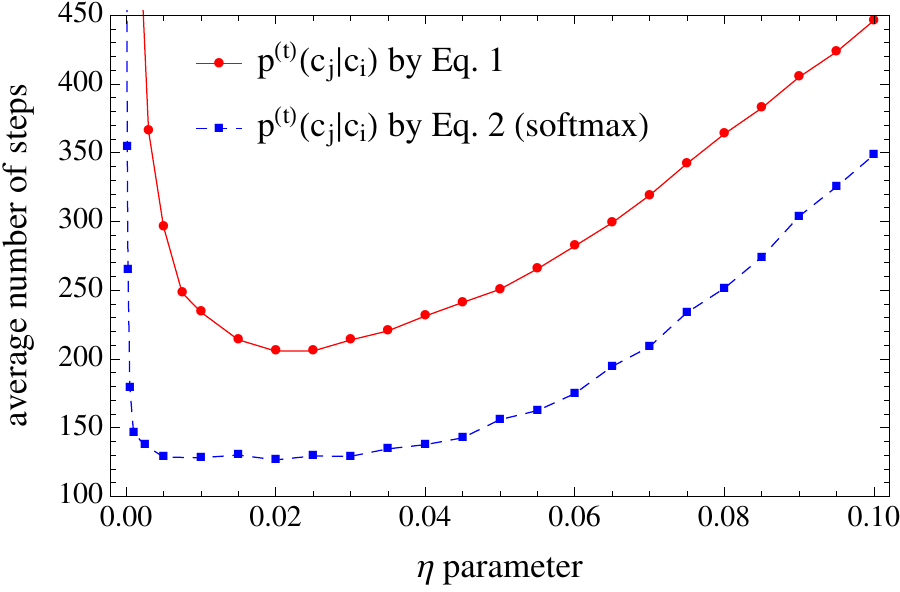}
        \label{fig:MCrandomEta}
    }
    \subfigure[]
    {
        \includegraphics[width=7cm]{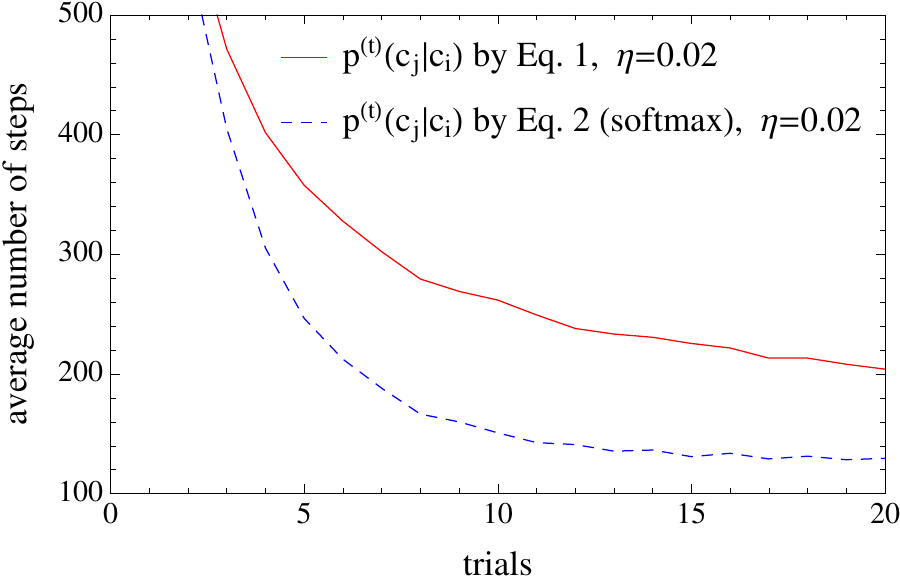}
        \label{fig:MCrandomLearning}
    }
    \caption{Performance of the PS agent in the mountain-car problem during 20 trials. Initially the agent has a random position and velocity. Solid red curves depict the PS performance using the basic transition probability function (Eq.~(\ref{eq:probablititesBasic})). Dashed blue curves depict the PS performance using the softmax transition probability function (Eq.~(\ref{eq:probablititesBoltzmann})). A damping value of $\gamma=0$ is used throughout. All curves are averaged over $10^4$ agents. 
(a) The dependence of the PS performance on the $\eta$ parameter is shown after 20 trials. (b) PS learning curves are shown for optimal values of $\eta=0.02$ (for 20 trials). 
The performance improves with the number of trials: from about 735 steps at the first trial to 204 (solid red) and 129 (dashed blue) steps, after 20 trials.
}
    \label{fig:MCrandom}
\end{figure}

To examine the PS model in the mountain-car problem, we simulate its performance with the two-layered clip network depicted in Fig.~\ref{fig:memoryMC}, where  
the continuous input space is discretized uniformly. Specifically, we chose a grid of 20 by 20 for a later comparison with existing results from the literature (see below). As before, we study the model with both choices for the transition probability given in Eqs.~(\ref{eq:probablititesBasic}) and (\ref{eq:probablititesBoltzmann}), and use the generalized update rules of Eq.~(\ref{eq:hupdate2})-(\ref{eq:gupdate}). To get an optimal performance we set the damping parameter to $\gamma=0$, as the environment does not change. To find an optimal $\eta$ parameter we look at the PS performance after 20 trials as a function of $\eta$, as shown in Fig.~\ref{fig:MCrandomEta}. Both transition function of Eqs.~(\ref{eq:probablititesBasic}) and (\ref{eq:probablititesBoltzmann}) are shown in solid red and dashed blue curves, respectively. It is seen that in both cases, an optimal $\eta$ value exists at $0.02$. In general, it is seen that the dependence on $\eta$ resembles the one observed in the grid-world task (see Fig.~\ref{fig:GridEta}). In particular, the PS performance is quite bad for $\eta \rightarrow 0$ and $\eta \rightarrow 1$ for the same reasons described in Sec.~\ref{sec:GridWorld}. 
In addition, we see here too that the softmax function of Eq.~(\ref{eq:probablititesBoltzmann}) improves the obtained performances and results with a performance curve that is less sensitive to changes in the $\eta$ parameter.  

Fig.~\ref{fig:MCrandomLearning} shows the learning curve of the PS agent using the optimal value $\eta=0.02$, as found above. The average number of steps required to reach the goal is shown for each trial for both the basic and the softmax function in solid red and dashed blue curves, respectively. It is seen that the PS agents manage to find the goal with less steps when the number of trials increases, as required. As in the grid-world task, the softmax function for the probability function improves not only the final performance but also the rate of the learning. For later comparison we indicate that with the softmax function the agents make about 223 steps per trial, averaged over the first 20 trials.

For comparison, we next look at the performance of the SARSA algorithm~\cite{rummery1994line} in the mountain-car problem, as reported in~\cite{singh1996reinforcement}. For completeness we shortly note that the SARSA algorithm estimates an ``action-value" function which gives an expected future reward for any percept-action pair. At each time step the action that maximizes this future reward is deterministically chosen. In Ref.~\cite{singh1996reinforcement} the infinite input space of the mountain-car problem was represented by five $9$ by $9$ grids, each of which is offset by a random fraction of a one cell's width. Each of the five grids was associated with its own action-value functions and an action was chosen according to the largest sum over the corresponding values from all grids. Here a reward of $-1$ was given at all time steps, except when reaching the goal, or as stated by the authors ``passing the top ends the trial and ends this punishment". Compared with our rewarding scheme there is thus a constant shift of rewards by $-1$. The reported performance of the SARSA algorithm is 450 steps per trial, averaged over the first 20 trials. 
According to the above results of the PS, we note that the PS is twice as fast (with 223 steps per trial). 

\begin{figure}[h!]
    \centering
    \subfigure[]
    {
        \includegraphics[width=7cm]{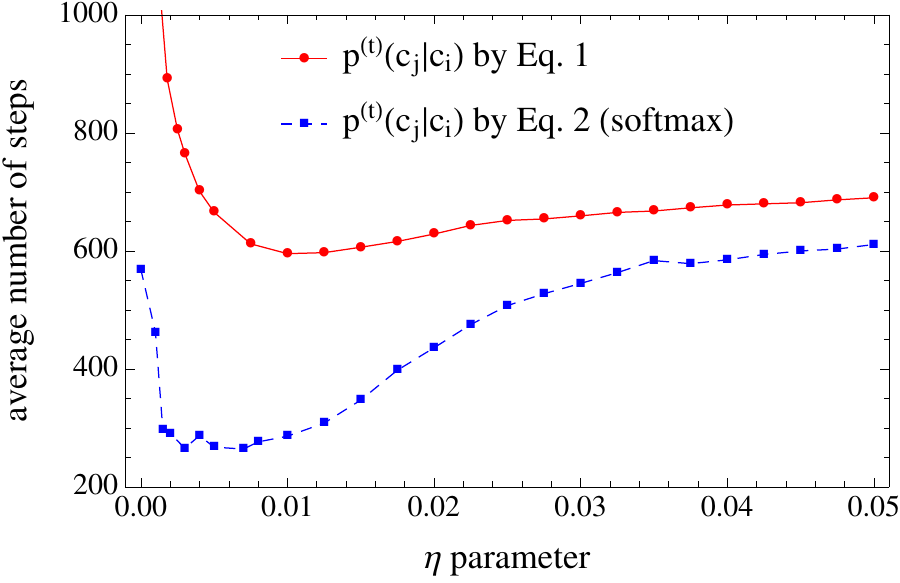}
        \label{fig:MCbottomEta}
    }
    \subfigure[]
    {
        \includegraphics[width=7cm]{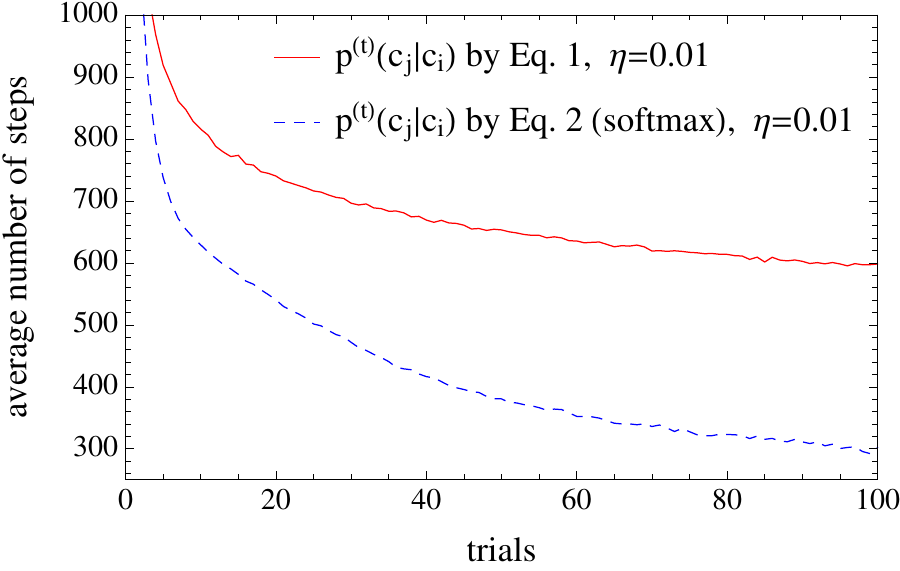}
        \label{fig:MCbottomLearning}
    }
    \caption{Performance of the PS agent in the mountain-car problem during 100 trials. Initially the agent is placed at $x=-0.5$ with zero velocity. Solid red curves depict the PS performance using the basic transition probability function (Eq.~(\ref{eq:probablititesBasic})). Dashed blue curves depict the PS performance using the softmax transition probability function (Eq.~(\ref{eq:probablititesBoltzmann})). A damping value of $\gamma=0$ is used throughout. All curves are averaged over $10^4$ agents. 
(a) The dependence of the PS performance on the $\eta$ parameter is shown after 100 trials. (b) PS learning curves are shown for optimal values of $\eta=0.01$ (for 100 trials). 
The performance improves with the number of trials: from about 1450 steps at the first trial to 593 (solid red) and 302 (dashed blue) steps, after 100 trials.}
    \label{fig:MCbottom}
\end{figure}

Comparing the PS agent with a more recent implementation of the SARSA algorithm for the mountain-car problem \cite{sutton2012dyna} we next consider the case in which the agent has a fixed initial position and velocity (as opposed to random ones). Following Ref.~\cite{sutton2012dyna}, we set the agent to $x=-0.5$ and $v=0$ at the beginning of each trial, i.e.\ almost at the bottom of the hill with zero velocity, and checked its performance after 100 trials. We discretized the input space to a uniform 20 by 20 grid, as before, and chose an optimal glow parameter $\eta=0.01$ according to the agent's best performance after 100 trials (see Fig.~\ref{fig:MCbottomEta}) using both the basic probability transition functions (solid red curve) and the softmax function (dashed blue curve). 
The corresponding learning curves are displayed in Fig.~\ref{fig:MCbottomLearning}, where it is shown that the PS agent manages to learn and to reach the goal with a decreasing number of steps in this scenario too. This fixed initial starting point turns out, however, to be rather difficult for the agent as it should first drive away from the goal (see Appendix \ref{sec:Appendix}). Empirically, we see that even after 100 trials the PS agent (with the softmax distribution) requires as many as 302 steps to find the goal, more steps than it needs in the case of random initial coordinates, after 20 trials alone. 
 
In what follows we compare the performances of the PS agent with those of the SARSA algorithm as presented in Fig.~3 of Ref.~\cite{sutton2012dyna}. In this reference, a reward of $-1$ was given at each time step until the goal was found, an action was chosen according to an $\epsilon$-greedy policy with $\epsilon=0.1$, and the 
value function was represented with 10 grids, each of $10^4$ input states, using  a linear function approximation. It is seen in Fig.~3 of Ref.~\cite{sutton2012dyna} that after 100 trials the SARSA algorithm is able to find the goal in about $150$ steps, i.e.\ twice as fast as the PS agent with the softmax transition function. We relate the relative success of SARSA in this case to the combined usage of a dense grid discretization (of $10^4$ states), a large number of grids (10), and a function approximation for the value function~\cite{sutton2012dyna}. As described above, the PS implementation for the mountain-car task is more economic: each agent is supplied with only a single network of 400 percept clips, where no kind of function approximation is used (implying that each percept-action edge has to be learned independently). This can be improved. For example, with 900 percepts the PS performance (using the softmax function) already improves by $\approx10\%$ and the agents find the goal after 276 steps on average. 

Table~\ref{table:comparison} summarizes the performances of the PS model in the mountain-car task, in both cases of random and fixed initial position. The performances of the SARSA algorithm in the corresponding scenarios are also shown, for comparison.

\begin{table}[h!]
\begin{tabular}{c|c|c|c|c}
 Initial state & \# of trials & Model & Performance & Parameters  \\[1mm] 
 \hline Random $ x $ and $ v $ & 20 & PS  & 313/trial & 20 by 20 input space \\
  & & PS (softmax) & 223/trial & $ \lambda = 1 $, $ \eta = 0.02 $, $ \gamma = 0 $ \\ \cline{3-5}
  & & SARSA~\cite{singh1996reinforcement}  & 450/trial & 5 grids, each of 9 by 9 input space\\
 \hline $ x = -0.5 $, $ v = 0 $ & 100 & PS & 593 & 20 by 20 input space \\
  & & PS (softmax) & 302 & $ \lambda = 1 $, $ \eta = 0.01 $, $ \gamma = 0 $\\ \cline{3-5}
  & & SARSA~\cite{sutton2012dyna} & 150 & 10 grids, \\
  & & & & each with $10^4$ input space \\
\end{tabular}
\caption{Mountain car: performances of the PS model in comparison with the SARSA algorithm, as reported in \cite{singh1996reinforcement, sutton2012dyna}.}.
\label{table:comparison}
\end{table}

\section{Conclusion}
\label{sec:Conclusion}

We studied the performance of the projective simulation (PS) agent in the navigation tasks of grid-world and mountain-car, in which an agent is supposed to learn how to find a goal in minimal number of steps. In the grid-world task the agent has to deal with a delayed reward that is given only at the end of the trial and with a large input space. In particular, there exists many (infinite) different paths to the goal, but only a few of them are optimal. In the mountain-car task the input state space is even infinite, adding the challenge of how to learn with only finite number of possible input percepts. 

In both tasks we saw that the PS agent manages to find the goal faster after each trial. The PS agent starts the first trial by randomly trying available actions until it accidentally reaches the goal. 
On average, during the first trial the PS finds the goal after 870 steps in the grid-world task, and after 735 steps (1450 steps when the initial coordinates are fixed) in the mountain-car task. With appropriately chosen damping parameters the PS greatly improves its performance: 
In the grid-world task the number of steps to reach the goal goes down to 15.4 after 100 trials; In the mountain-car task, the number of steps to reach the goal goes down to 129 steps after 20 trials using randomized initial coordinates, and to 302 steps after 100 trials using fixed initial coordinates.
These results were obtained using the softmax transition probability function of Eq.~(\ref{eq:probablititesBoltzmann}) which, due to rescaling of the hopping probability, always improves the performance compared to the use of the basic transition probability function of Eq.~(\ref{eq:probablititesBasic}). 
We further studied the performance of the PS as a function of the glow parameter $\eta$ and showed that the edge-glow mechanism of the PS plays an important role in scenarios where the reward is delayed.

The performance of the PS agent was compared with those of the policy iteration (PI) and the SARSA algorithms. Qualitatively, the performance of the PS model is comparable to those of the other models, and no major differences were observed. Quantitatively, we saw that in the mountain-car, when starting from a fixed coordinate, PS does not perform as good as SARSA. We showed, however, that to a certain extent one can improve the PS performance by increasing the input state space, i.e.\ the number of possible percept clips. We further showed that the PS agent performs almost as good as the PI agent in the grid-world task and that it outperforms the SARSA algorithm in the mountain-car task when the initial coordinates are chosen randomly. We thus conclude that the PS model performs well also in navigation scenarios with large and even continuous input space.\footnote{Even though it is not the topic of this paper, it should be mentioned that the PS model has the additional benefit that it can be straightforwardly generalized to quantum mechanical operation~\cite{briegel2012projective, paparo2014quantum}. An agent based on quantum projective simulation (Q-PS) has recently been shown to provide a significant speed-up for active learning scenarios~\cite{paparo2014quantum}.}

\appendix
\section{Analyzing the physics background of the mountain-car problem}
\label{sec:Appendix}

\begin{figure}[h!]
   \begin{center}
   \begin{tabular}{c}
   \includegraphics[height=5cm]{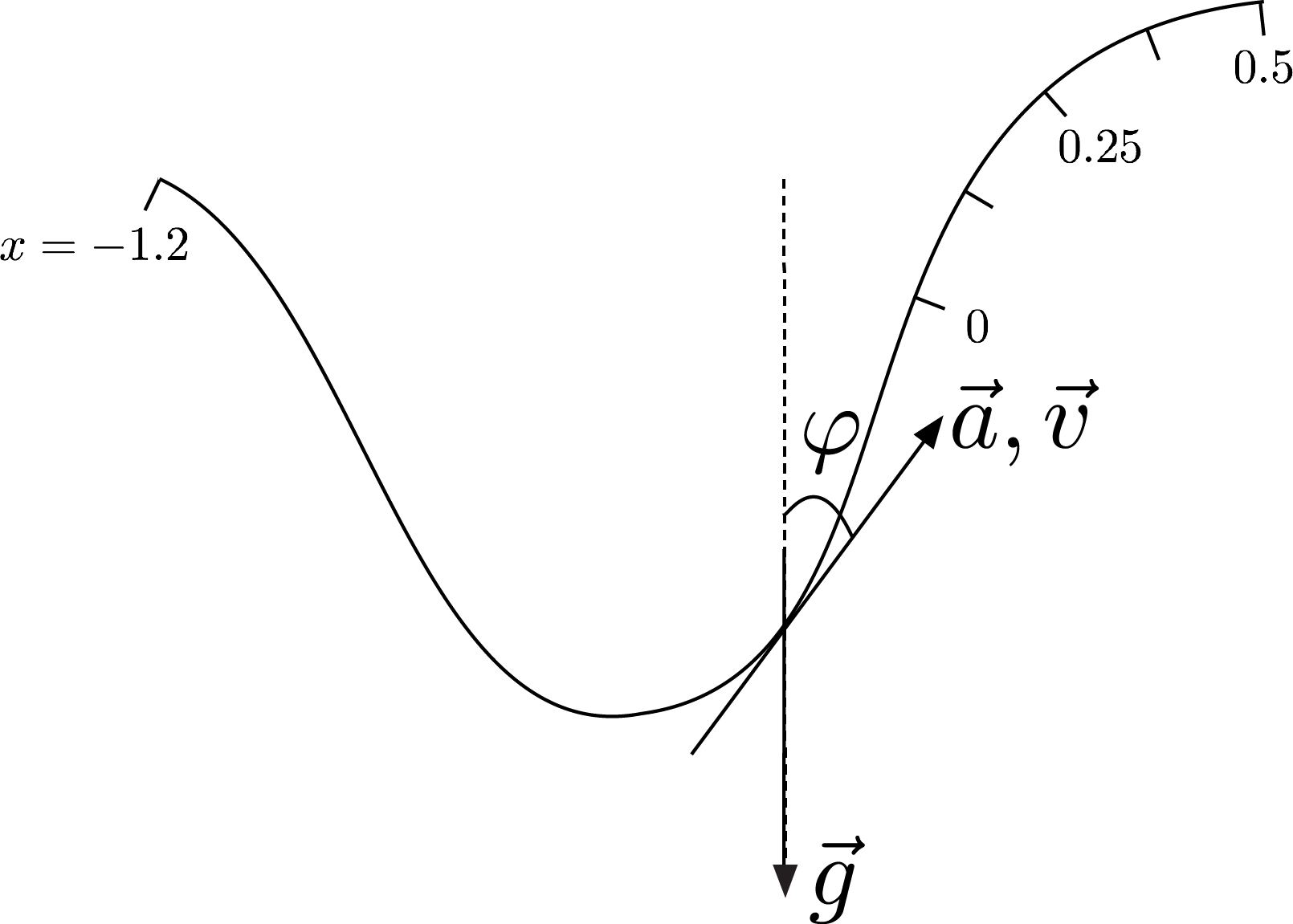}
   \end{tabular}
   \end{center}
\caption{Accelerations and velocity in the mountain-car problem. The marks on the road represent $x$ coordinate.}
\label{fig:MCanalytics}
   \end{figure}

A car of mass $m$ drives up the hill with a tangential velocity $\vec{v}$, it has an engine acceleration $\vec{a}$ and experiences a gravitational acceleration $\vec{g}$ as illustrated in Fig.~\ref{fig:MCanalytics}. Its equation of motion is therefore given by:
\begin{equation}
  d\vec{v}/dt = \vec{a} + \vec{g}.
\end{equation}
By projecting the vectors onto the direction of motion we get
\begin{equation}
  \begin{array}{ccl}
    dv/dt & = & a - g \cos\varphi \\
    dx/dt & = & v_x = v,
  \end{array}
\end{equation}
where $\varphi$ is the angle between the vectors $\vec{a}$ and $-\vec{g}$. Integration leads to 
\begin{equation}
  \begin{array}{ccl}
    v_t & = & v_{t-1} + \int_{t-1}^t a_\tau d\tau - g \int_{t-1}^t \cos\varphi_\tau d\tau \\
    x_t & = & x_{t-1} + \int_{t-1}^t v_\tau d\tau.
  \end{array}
  \label{eq:phStateUpdate}
\end{equation}
We next assume that during the small time interval $dt=1$ the integrands $a_\tau$, $\cos\varphi_{\tau}$ and $v_{\tau}$ do not change much, thus we obtain the following approximations for the $(x_{t+1},v_{t+1})$ coordinates of the next time step: 
\begin{equation}
  \begin{array}{ccl}
    v_t & \approx & v_{t-1} + a_{t-1} - g \cos(\varphi_{t-1}) \\
    x_t & \approx & x_{t-1} + v_{t-1}.
  \end{array}
\label{eq:phStateApproxUpdate}
\end{equation}

\begin{figure}[h!]
   \begin{center}
   \begin{tabular}{c}
   \includegraphics[height=5cm]{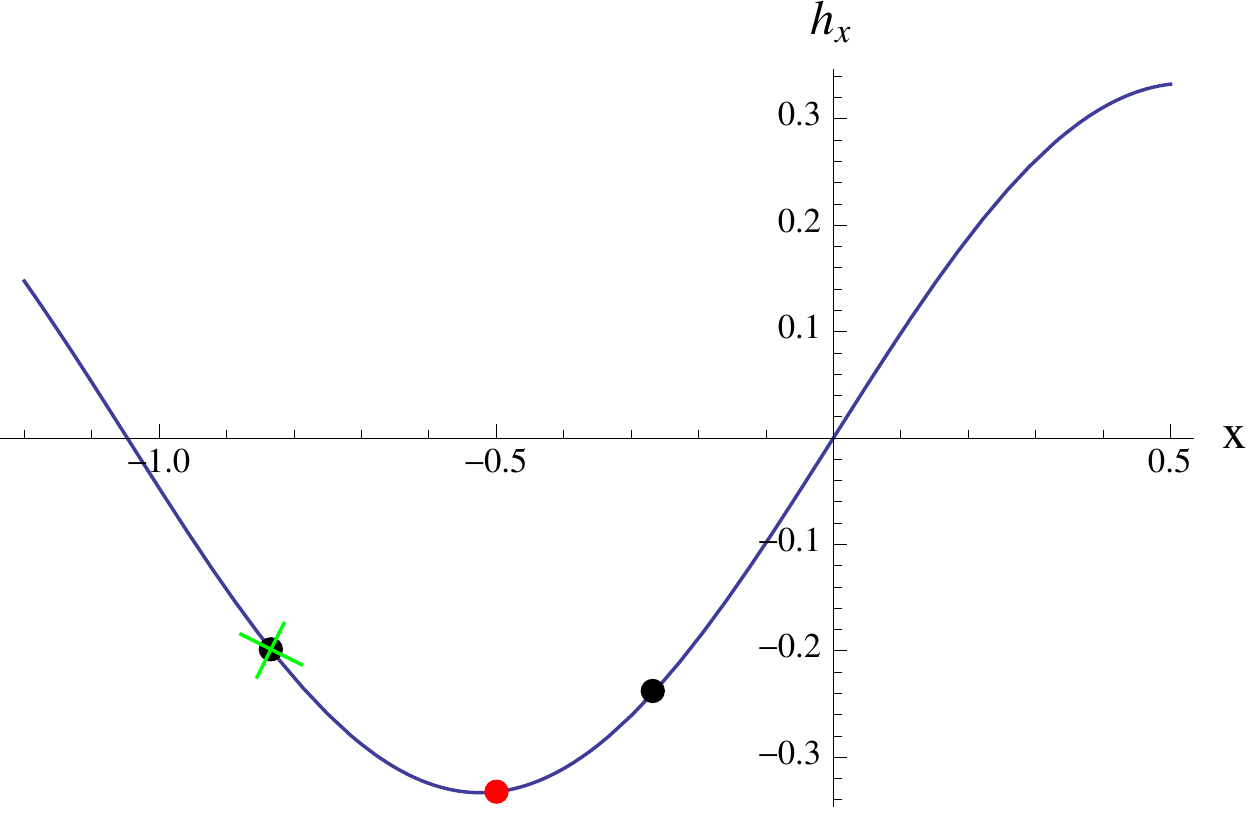}
   \end{tabular}
   \end{center}
\caption{The height of the hill in the mountain-car task for $g=0.0025$ and $C = 0$. Red point shows the initial position. Two black point are at $x \approx -0.834$ and $x \approx -0.27$, the green cross is a little bit lower ($x \approx -0.832$) than the left black one.}
\label{fig:MCpotential}
   \end{figure}

Comparing Eq.~(\ref{eq:phStateApproxUpdate}) with Eq.~(\ref{eq:stateUpdate}) one obtains $a_{t-1}= 0.001*\mathrm{Action} $ and $ g \cos\varphi_{t-1} = 0.0025\cos (3x_{t-1}) $. 
A change of the height is given by $dh = dx \cos \varphi_{x}$, which integrates to
\begin{equation}
h_x = \frac{0.0025}{g}\int \cos (3x) dx = \frac{0.0025}{3g}\sin(3x)+C.
\label{eq:amplitude}
\end{equation}
The height $h_x$ is plotted in~Fig.~\ref{fig:MCpotential} for g=0.0025 (using the same convention as in \cite{singh1996reinforcement}) and $C=0$ , where $x$ is the coordinate perceived by an agent, e.g. marks on the road. Eq.~(\ref{eq:amplitude}) indicates that the bottom of the hill is placed at $x=-\pi/6$.  

If the agent starts near the bottom of the hill at $x_0 = -0.5$ with a zero velocity $v_0=0$ and pushes the car in the direction towards the top of the mountain, its engine power will not be sufficient. To see that we equate the net work done by an engine with the difference in total energy (potential plus kinetic) 
\begin{equation}
 ma(x_\mathrm{goal}-x_0) = mg (h_\mathrm{goal} - h_0) + \frac{mv_f^2}{2} 
\label{eq:work}
\end{equation}
from which we get
\begin{equation}
 a(x_\mathrm{goal}-x_0) \geq g (h_\mathrm{goal} - h_0) \Rightarrow x_\mathrm{goal} \geq -0.5 + \frac{5}{6}\Big(\sin (3x_\mathrm{goal}) - \sin(-1.5)\Big)
\end{equation}
which does not hold for $x_\mathrm{goal}=0.5$.

To get the maximum $x$ value the agent can reach by always pushing right we use Eq.~(\ref{eq:work}) and set the final velocity $v_f=0$, which gives $x_{max}\approx-0.27$ (which is indeed much before the goal at $x=0.5$). A similar analysis shows that by pushing the car always to the left the agent could reach $x\approx-0.834$. Fig.~\ref{fig:MCpotential} marks in black circles the furthest points the agent can reach by pushing the car only in one direction (their hight differ because the initial position $-0.5$ is not exactly at the bottom of the hill). An additional green cross marks the point from which the agent could reach the goal by just pushing the car to the right. One possible strategy is to go from $x=-0.5$ to the left till the car stops at $x\approx -0.834$ and from that point on always push the car in the direction of the goal. With this strategy one can reach the goal in 89 steps: 36 actions to the left, followed by 63 actions to the right. Note that in this analysis we did not enforce explicitly the bounds on the velocity. This is, however, unnecessary as within the above particular strategy the absolute value of the velocity during the whole trial is always less than the maximum allowed value of $0.07$.

 \newcommand{\noop}[1]{}


\begin{thebibliography}{20}
\expandafter\ifx\csname natexlab\endcsname\relax\def\natexlab#1{#1}\fi
\expandafter\ifx\csname bibnamefont\endcsname\relax
  \def\bibnamefont#1{#1}\fi
\expandafter\ifx\csname bibfnamefont\endcsname\relax
  \def\bibfnamefont#1{#1}\fi
\expandafter\ifx\csname citenamefont\endcsname\relax
  \def\citenamefont#1{#1}\fi
\expandafter\ifx\csname url\endcsname\relax
  \def\url#1{\texttt{#1}}\fi
\expandafter\ifx\csname urlprefix\endcsname\relax\def\urlprefix{URL }\fi
\providecommand{\bibinfo}[2]{#2}
\providecommand{\eprint}[2][]{\url{#2}}

\bibitem[{\citenamefont{Briegel and De~las
  Cuevas}(2012)}]{briegel2012projective}
\bibinfo{author}{\bibfnamefont{H.~J.} \bibnamefont{Briegel}} \bibnamefont{and}
  \bibinfo{author}{\bibfnamefont{G.}~\bibnamefont{De~las Cuevas}},
  \bibinfo{journal}{Scientific reports} \textbf{\bibinfo{volume}{2}}
  (\bibinfo{year}{2012}).

\bibitem[{\citenamefont{Russell et~al.}(1995)\citenamefont{Russell, Norvig,
  Canny, Malik, and Edwards}}]{russell1995artificial}
\bibinfo{author}{\bibfnamefont{S.~J.} \bibnamefont{Russell}},
  \bibinfo{author}{\bibfnamefont{P.}~\bibnamefont{Norvig}},
  \bibinfo{author}{\bibfnamefont{J.~F.} \bibnamefont{Canny}},
  \bibinfo{author}{\bibfnamefont{J.~M.} \bibnamefont{Malik}}, \bibnamefont{and}
  \bibinfo{author}{\bibfnamefont{D.~D.} \bibnamefont{Edwards}},
  \emph{\bibinfo{title}{Artificial intelligence: a modern approach}},
  vol.~\bibinfo{volume}{2} (\bibinfo{publisher}{Prentice Hall Englewood
  Cliffs}, \bibinfo{year}{1995}).

\bibitem[{\citenamefont{Sutton}(1998)}]{barto1998reinforcement}
\bibinfo{author}{\bibfnamefont{A.~G.} \bibnamefont{Sutton}, \bibfnamefont{R.~S.
  \&~Barto}}, \emph{\bibinfo{title}{Reinforcement learning: An introduction}}
  (\bibinfo{publisher}{MIT press}, \bibinfo{year}{1998}).

\bibitem[{\citenamefont{Wiering and van Otterlo~(Eds.)}(2012)}]{RL_2012_book}
\bibinfo{author}{\bibfnamefont{M.}~\bibnamefont{Wiering}} \bibnamefont{and}
  \bibinfo{author}{\bibfnamefont{M.}~\bibnamefont{van Otterlo~(Eds.)}},
  \emph{\bibinfo{title}{Reinforcement learning: State of the Art}},
  vol.~\bibinfo{volume}{12} (\bibinfo{publisher}{Springer},
  \bibinfo{year}{2012}).

\bibitem[{\citenamefont{Mautner et~al.}(2014)\citenamefont{Mautner, Makmal,
  Manzano, Tiersch, and Briegel}}]{mautner2013projective}
\bibinfo{author}{\bibfnamefont{J.}~\bibnamefont{Mautner}},
  \bibinfo{author}{\bibfnamefont{A.}~\bibnamefont{Makmal}},
  \bibinfo{author}{\bibfnamefont{D.}~\bibnamefont{Manzano}},
  \bibinfo{author}{\bibfnamefont{M.}~\bibnamefont{Tiersch}}, \bibnamefont{and}
  \bibinfo{author}{\bibfnamefont{H.~J.} \bibnamefont{Briegel}},
  \bibinfo{journal}{New Generation Computing, in Press, arXiv:1305.1578}
  (\bibinfo{year}{2014}).

\bibitem[{\citenamefont{Wilson}(1995)}]{Wilson95}
\bibinfo{author}{\bibfnamefont{S.~W.} \bibnamefont{Wilson}},
  \bibinfo{journal}{Evolutionary computation} \textbf{\bibinfo{volume}{3}},
  \bibinfo{pages}{149} (\bibinfo{year}{1995}).

\bibitem[{\citenamefont{Sutton}(1990)}]{sutton1990integrated}
\bibinfo{author}{\bibfnamefont{R.~S.} \bibnamefont{Sutton}}, in
  \emph{\bibinfo{booktitle}{Proceedings of the 7th International Conference on
  Machine Learning}} (\bibinfo{year}{1990}), pp. \bibinfo{pages}{216--224}.

\bibitem[{\citenamefont{Moore}(1990)}]{moore1990efficient}
\bibinfo{author}{\bibfnamefont{A.~W.} \bibnamefont{Moore}}, Ph.D. thesis,
  \bibinfo{school}{University of Cambridge} (\bibinfo{year}{1990}).

\bibitem[{\citenamefont{Crook and Hayes}(2003)}]{crook2003learning}
\bibinfo{author}{\bibfnamefont{P.}~\bibnamefont{Crook}} \bibnamefont{and}
  \bibinfo{author}{\bibfnamefont{G.}~\bibnamefont{Hayes}},
  \bibinfo{journal}{Towards Intelligent Mobile Robots}
  \textbf{\bibinfo{volume}{4}} (\bibinfo{year}{2003}).

\bibitem[{\citenamefont{Wiering and van
  Otterlo}(2012)}]{wiering2012reinforcement}
\bibinfo{author}{\bibfnamefont{M.}~\bibnamefont{Wiering}} \bibnamefont{and}
  \bibinfo{author}{\bibfnamefont{M.}~\bibnamefont{van Otterlo}},
  \emph{\bibinfo{title}{Reinforcement Learning: State-of-the-art}},
  vol.~\bibinfo{volume}{12} (\bibinfo{publisher}{Springer},
  \bibinfo{year}{2012}).

\bibitem[{\citenamefont{Singh and Sutton}(1996)}]{singh1996reinforcement}
\bibinfo{author}{\bibfnamefont{S.~P.} \bibnamefont{Singh}} \bibnamefont{and}
  \bibinfo{author}{\bibfnamefont{R.~S.} \bibnamefont{Sutton}},
  \bibinfo{journal}{Machine learning} \textbf{\bibinfo{volume}{22}},
  \bibinfo{pages}{123} (\bibinfo{year}{1996}).

\bibitem[{\citenamefont{Sutton}(1996)}]{sutton1996generalization}
\bibinfo{author}{\bibfnamefont{R.~S.} \bibnamefont{Sutton}},
  \bibinfo{journal}{Advances in neural information processing systems}
  \textbf{\bibinfo{volume}{8}}, \bibinfo{pages}{1038} (\bibinfo{year}{1996}).

\bibitem[{\citenamefont{Smart and Kaelbling}(2000)}]{smart2000practical}
\bibinfo{author}{\bibfnamefont{W.~D.} \bibnamefont{Smart}} \bibnamefont{and}
  \bibinfo{author}{\bibfnamefont{L.~P.} \bibnamefont{Kaelbling}}, in
  \emph{\bibinfo{booktitle}{Proceedings of the International Conference on
  Machine Learning}} (\bibinfo{organization}{Citeseer}, \bibinfo{year}{2000}),
  pp. \bibinfo{pages}{903--910}.

\bibitem[{\citenamefont{Rasmussen et~al.}(2003)\citenamefont{Rasmussen, Kuss
  et~al.}}]{rasmussen2003gaussian}
\bibinfo{author}{\bibfnamefont{C.~E.} \bibnamefont{Rasmussen}},
  \bibinfo{author}{\bibfnamefont{M.}~\bibnamefont{Kuss}}, \bibnamefont{et~al.},
  in \emph{\bibinfo{booktitle}{Proceedings of the Conference on Neural
  Information Processing Systems}} (\bibinfo{year}{2003}).

\bibitem[{\citenamefont{Jong and Stone}(2006)}]{jong2006kernel}
\bibinfo{author}{\bibfnamefont{N.}~\bibnamefont{Jong}} \bibnamefont{and}
  \bibinfo{author}{\bibfnamefont{P.}~\bibnamefont{Stone}}, in
  \emph{\bibinfo{booktitle}{Proceedings of the International Conference on
  Machine Learning}} (\bibinfo{year}{2006}).

\bibitem[{\citenamefont{Whiteson and Stone}(2006)}]{whiteson2006evolutionary}
\bibinfo{author}{\bibfnamefont{S.}~\bibnamefont{Whiteson}} \bibnamefont{and}
  \bibinfo{author}{\bibfnamefont{P.}~\bibnamefont{Stone}},
  \bibinfo{journal}{The Journal of Machine Learning Research}
  \textbf{\bibinfo{volume}{7}}, \bibinfo{pages}{877} (\bibinfo{year}{2006}).

\bibitem[{\citenamefont{Heidrich-Meisner and
  Igel}(2008)}]{heidrich2008variable}
\bibinfo{author}{\bibfnamefont{V.}~\bibnamefont{Heidrich-Meisner}}
  \bibnamefont{and} \bibinfo{author}{\bibfnamefont{C.}~\bibnamefont{Igel}}, in
  \emph{\bibinfo{booktitle}{Recent Advances in Reinforcement Learning}}
  (\bibinfo{publisher}{Springer}, \bibinfo{year}{2008}), pp.
  \bibinfo{pages}{136--150}.

\bibitem[{\citenamefont{Sutton et~al.}(2012)\citenamefont{Sutton,
  Szepesv{\'a}ri, Geramifard, and Bowling}}]{sutton2012dyna}
\bibinfo{author}{\bibfnamefont{R.~S.} \bibnamefont{Sutton}},
  \bibinfo{author}{\bibfnamefont{C.}~\bibnamefont{Szepesv{\'a}ri}},
  \bibinfo{author}{\bibfnamefont{A.}~\bibnamefont{Geramifard}},
  \bibnamefont{and} \bibinfo{author}{\bibfnamefont{M.~P.}
  \bibnamefont{Bowling}}, \bibinfo{journal}{arXiv:1206.3285}
  (\bibinfo{year}{2012}).

\bibitem[{\citenamefont{Rummery and Niranjan}(1994)}]{rummery1994line}
\bibinfo{author}{\bibfnamefont{G.~A.} \bibnamefont{Rummery}} \bibnamefont{and}
  \bibinfo{author}{\bibfnamefont{M.}~\bibnamefont{Niranjan}},
  \emph{\bibinfo{title}{On-line Q-learning using connectionist systems}}
  (\bibinfo{publisher}{University of Cambridge}, \bibinfo{year}{1994}).

\bibitem[{\citenamefont{Paparo et~al.}(2014)\citenamefont{Paparo, Dunjko,
  Makmal, Martin-Delgado, and Briegel}}]{paparo2014quantum}
\bibinfo{author}{\bibfnamefont{G.~D.} \bibnamefont{Paparo}},
  \bibinfo{author}{\bibfnamefont{V.}~\bibnamefont{Dunjko}},
  \bibinfo{author}{\bibfnamefont{A.}~\bibnamefont{Makmal}},
  \bibinfo{author}{\bibfnamefont{M.~A.} \bibnamefont{Martin-Delgado}},
  \bibnamefont{and} \bibinfo{author}{\bibfnamefont{H.~J.}
  \bibnamefont{Briegel}}, \bibinfo{journal}{submitted to Phys. Rev. X, arXiv:1401.4997}
  (\bibinfo{year}{2014}).

\end{thebibliography}
\end{document}